\documentclass[10pt,twocolumn,letterpaper]{article}

\usepackage{cvpr}              

\usepackage{graphicx}
\usepackage{amsmath}
\usepackage{amssymb}
\usepackage{booktabs}
\usepackage{bbm}
\usepackage[table,xcdraw]{xcolor}
\usepackage[pagebackref,breaklinks,colorlinks]{hyperref}
\usepackage[ruled,vlined, linesnumbered]{algorithm2e}
\usepackage{amsmath,amsfonts,bm}
\usepackage{algorithm2e}
\usepackage[parfill]{parskip}

\usepackage[toc]{appendix}
\usepackage{minitoc}

\usepackage{sidecap}
\usepackage{floatrow}
\usepackage{sidecap}
\usepackage[utf8]{inputenc} 
\usepackage[T1]{fontenc}    
\usepackage{url}            
\usepackage{booktabs}       
\usepackage{amsfonts}       
\usepackage{nicefrac}       
\usepackage{microtype}      
\usepackage{lipsum}		
\usepackage{graphicx}
\usepackage{doi}

\newcommand{\G}{\mathcal{G}}

\newcommand{\U}{\mathcal{U}}

\newcommand{\FID}{$\text{F}^2\text{ID}$}
\newcommand{\Real}{\mathbb{R}}

\newcommand{\F}{\mathcal F}

\newcommand{\A}{\mathcal A}
\newcommand{\M}{\mathcal M}
\renewcommand{\P}{\mathcal P}

\newcommand{\D}{\mathcal D}

\newcommand{\V}{\mathcal V}

\newcommand{\C}{\mathcal C}

\newcommand{\E}{\mathbb E}

\newcommand{\Prob}{\mathbb P}

\newcommand{\bG}{\boldsymbol{G}}
\newcommand{\bL}{\boldsymbol{L}}

\newtheorem{proof*}{Proof}[section]

\newtheorem{definition*}{Definition}[section]

\usepackage[capitalize]{cleveref}
\crefname{section}{Sec.}{Secs.}
\Crefname{section}{Section}{Sections}
\Crefname{table}{Table}{Tables}
\crefname{table}{Tab.}{Tabs.}

\def\cvprPaperID{*****} 
\def\confName{CVPR}
\def\confYear{2023}

\begin{document}

\title{\textit{PaCMO:} Partner Dependent Human Motion Generation in Dyadic Human Activity using Neural Operators}

\author{Md Ashiqur Rahman\textsuperscript{\rm 1}, Jasorsi Ghosh\textsuperscript{\rm 1}, Hrishikesh Viswanath\textsuperscript{\rm 1}, Kamyar Azizzadenesheli\textsuperscript{\rm 2}, Aniket Bera\textsuperscript{\rm 1}\\
\textsuperscript{\rm 1}Department of Computer Science, Purdue University, West Lafayette, Indiana.\\
\textsuperscript{\rm 2}Nvidia Corporation, Santa Clara, CA.\\
\{rahman79, ghosh117, hviswan\}@purdue.edu, kamyara@nvidia.com, ab@cs.purdue.edu}
\maketitle
\begin{abstract}
We address the problem of generating 3D human motions in dyadic activities. In contrast to the concurrent works, which mainly focus on generating the motion of a single actor from the textual description, we generate the motion of one of the actors from the motion of the other participating actor in the action. This is a particularly challenging, under-explored problem, that requires learning intricate relationships between the motion of two actors participating in an action and also identifying the action from the motion of one actor. To address these, we propose \textbf{pa}rtner \textbf{c}onditioned \textbf{m}otion \textbf{o}perator (PaCMO), a neural operator-based generative model which learns the distribution of human motion conditioned by the partner's motion in function spaces through adversarial training. Our model can handle long unlabeled action sequences at arbitrary time resolution. We also introduce  the "Functional Fréchet Inception Distance" $\text{F}^2\text{ID}$ metric for capturing similarity between real and generated data for function spaces. We test PaCMO on NTU RGB+D and DuetDance datasets and our model produces realistic results evidenced by $\text{F}^2\text{ID}$ score and the conducted user study.
\end{abstract} 
\section{Introduction}

Generating 3D human motion is widely studied and is a core component of animation, games, human-robot interaction, and AR/VR applications. Despite many efforts from the community, the problem of modeling 3D human motion is still poorly understood. Most of the works are focused on generating the motion of a single person \cite{temos,tevet2022motionclip,yan2019convolutional,ahuja2019language2pose}. These works often generate human motion without any interaction with the surrounding objects and rarely model multi-agent interactions. Modeling the motion of multiple people in an interactive setting is particularly difficult.

Recent works \cite{temos,tevet2022motionclip,song2022actformer, transVAE} primarily deal with generating motion of a single human with or without conditioned by action labels and sentences describing the action. The advancement of joint language-image models (such as CLIP) enables the production of motion sequences for unseen actions \cite{tevet2022motionclip, hong2022avatarclip}. But these models can only generate motion sequences of a fixed length at a specific time resolution. To synthesize lengthy (long duration) human motions many prior works assume an autoregressive or Markovian assumption \cite{ahn2018text2action,guo2020action2motion,zhang2020perpetual,lee2019dancing, ahuja2019language2pose}, which lead to a static state after a certain time. Many recent works \cite{temos,song2022actformer,transVAE,tevet2022motionclip, athanasiou2022teach} have utilized attention mechanisms \cite{vaswani2017attention} to capture non-local dependencies in long motion sequences. Though these models can generate long sequences but often in practice the produced sequences are of fixed length at a particular resolution. They can only generate human motion at a fixed time resolution which depends on the training data. But time continuous 3D human motion generation is crucial for animation and video games. At the same time, the problem of multi-person motion generation in interactive situations is under-explored. Though there have been attempts to address this problem in recent works \cite{song2022actformer, maheshwari2022mugl}, these methods require explicit action labels or sentences describing the action which may be impractical for AR/VR or HRI applications where we do not have explicit action labels. Rather, in many cases, we need to model the motion of digital humans interacting with real humans in arbitrary dyadic activities. In these situations, we need to generate the motion of a virtual human only from the motion of participating humans such that virtual humans can correctly collaborate with the intended activity of the human partner.

To address these problems, inspired by the recent development of generative modeling in functional spaces \cite{rahman2022generative}, we propose \textbf{PaCMO}, a conditional generative adversarial neural operator model to generate 3D human motion for dyadic activity from the movements of the other interacting human, i.e. the partner. To the best of our knowledge, PaCMO is the first to propose a time-continuous generative architecture to solve this problem. We model human motion sequence as a stochastic function of time. Given the motion of one of the actors in a dyadic activity, our model learns a push-forward measure from a Gaussian random field \cite{dudley2010sample} to the motion sequence of the other actor. Gaussian random field or Gaussian processes (GP) are continuous functions such that any finite collection of points on the function constitutes a multi-variate Gaussian distribution. Unlike traditional deep neural networks, neural operators are resolution invariant. As a result, our model can generate continuous time human motion and is not restricted to the resolution of the training data. And in evaluation time, PaCMO can generate motion at a resolution demanded by the application (e.g., matching require frame-rate for video game application).  PaCMO can efficiently model global dependencies in long sequences and is not limited by the receptive field of discrete convolution \cite{yan2019convolutional}. Also, PaCMO outputs the entire motion sequence at once, making it suitable for real-time applications and also avoiding the issue with autoregressive and Markovian models. 

Evaluating generative models is nontrivial and subject to human perception. Fréchet Inception Distance (FID) \cite{heusel2017gans} is a well used metric to quantitatively evaluate generative models. To obtain a closed-form solution of Fréchet distance, features from both generated and real data are modeled using multivariate Gaussian distributions. But finite dimensional multivariate Gaussian distributions cannot model features from infinite dimensional spaces. Also, such FID is not resolution invariant, as a fixed finite dimensional Gaussian distribution cannot model feature vectors of different sizes. As a result, FID is ill-suited to evaluate generative models in infinite dimensional spaces. To address this, we remove the finite-dimensional assumption and propose Functional FID or $\text{F}^2\text{ID}$, which is designed for general vector spaces and is resolution invariant.

In this work, we consider human motion as a sequence of skeletal poses but our model does not depend on any particular representation of human motion and can easily be extended to other representations, such as SMPL \cite{loper2015smpl}. We have evaluated our model on two widely used datasets: NTU-RGB+D-120 \cite{liu2019ntu} and DuetDance\cite{kundu2020duet} by both quantitative metrics and perceptual human evaluations. Our contributions can be summarized as follows:
\begin{itemize}
    \item We propose PaCMO, a novel deep neural operator for generating partner-dependent 3D human motion in interactive dyadic human activities without any action label or sentence describing the action.
    \item We introduce a novel conditional generative neural operator which is a resolution-invariant conditional generator for infinite dimensional spaces. 
    \item We propose ($\text{F}^2\text{ID}$), an evaluation metric for generative models in function spaces.
    \item Our model PaCMO can generate promising results outperforming baselines, taking the first step in partner-dependent human motion generation in an interactive setting. 
    
\end{itemize}
In the rest of the paper, in section \ref{sec:related work} we discuss related works for human motion generation and introduce neural operators to the readers. In section \ref{sec:method} we introduce the novel PaCMO architecture and ($\text{F}^2\text{ID}$) score. We discuss the details of our experiment with results in section \ref{sec:experiments}.
\section{Related Works}
\label{sec:related work}
\subsection{Human Motion Generation}
Human motion prediction and generation are well-studied problems in the fields of kinesiology, robotics, computer vision, and computer graphics. Especially the field of human motion generation has experienced rapid development following success in generative modeling such as GAN \cite{goodfellow2020gan}, VAE \cite{kingma2013auto}, normalizing flow \cite{kingma2018glow,papamakarios2021normalizing}, and diffusion \cite{dhariwal2021diffusion} based models. The task of human motion generation can be broadly divided into two categories : (a) conditional human motion generation, where the generation is guided by external variables such as action labels, action descriptions, or speech. (b) unconditional human motion generation, where we aim to model the entire space of human motion.

Earlier works in unconstrained motion generation primarily depend on autoregressive models \cite{habibie2017recurrent, yan2018mt, zhang2020perpetual}. CSGN \cite{yan2019convolutional} learns a mapping from latent vector sampled from the Gaussian process to human motion sequence by a series of convolution operations in an adversarial way. But due to the finite receptive field, it has the inherent limitation to capture very long-term dependencies and can only generate human motion at a fixed predetermined time resolution. NeMF \cite{he2022nemf} directly extends the recent developments in neural implicit represent action \cite{pumarola2021d, sitzmann2020implicit} to generative modeling but such models are extremely limited as the whole generated process which is infinite-dimensional (continuous human motion) depends only on a finite-dimensional latent vector.

Generating human motion from action label or description largely dominates the field on conditional motion generation \cite{guo2020action2motion,ahuja2019language2pose,ahn2018text2action}. Both ACTOR \cite{transVAE} and TEMOS \cite{temos} use a transformer architecture to generate human motion, whereas the former uses a pre-trained language model DistilBERT \cite{sanh2019distilbert} to encode the sentence description. GOAL \cite{taheri2022goal} generates human motion with the specific goal of grasping objects. TEACH \cite{athanasiou2022teach} is also a text-conditioned human motion generator but it provides temporal control over the generated motion. MotionCLIP \cite{tevet2022motionclip, hong2022avatarclip} leverages CLIP \cite{radford2021learning} to generate unseen actions and body shapes.

The problem of interactive multi-person motion generation is particularly challenging and requires joint modeling of multiple human motions. ActFormer \cite{song2022actformer} provides a framework to generate action-conditioned single and multi-person motion sequences using a transformer architecture. MUGL \cite{maheshwari2022mugl} uses a duration-aware autoregressive model to generate a multi-person motion sequence from an action label. 
\begin{figure*}[t]
    \centering
    \includegraphics[width=0.9\textwidth,trim={0cm 13cm 0cm 6cm},clip]{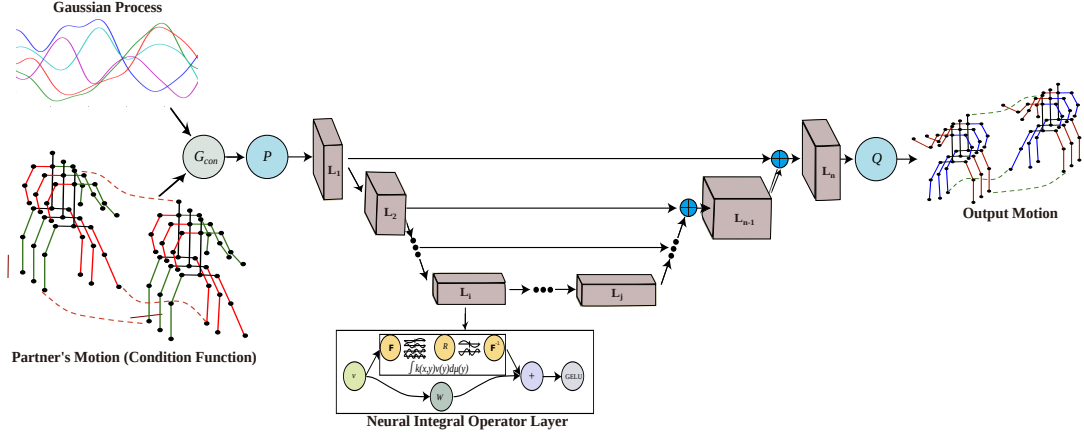}
    \caption{The PaCMO generator architecture. It takes the motion of one actor in a dyadic activity as the condition and generates the motion of the other partner.  The blue circles represent the concatenation operation in function spaces. }
    \label{fig:gen}
\end{figure*}
\begin{figure*}[t]
    \centering
    \includegraphics[width=0.9\textwidth,trim={0cm 0cm 0cm 1cm},clip]{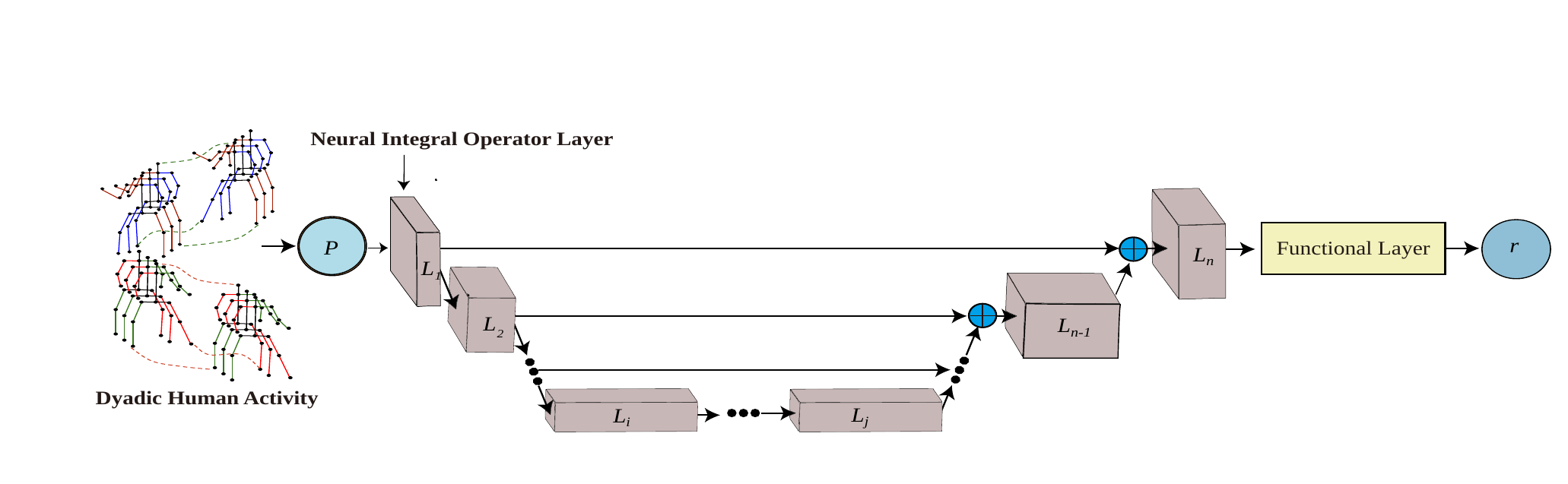}
    \caption{The PaCMO discriminator architecture. It takes the motion of two people involved in an interactive dyadic activity as input and maps it to a real number $r \in \Real$. }
    \label{fig:dis}
\end{figure*}

\subsection{Neural Operators}
Neural Operators \cite{li2020fourier,li2020neural,kovachki2021neural} are a new paradigm in machine learning that learns a mapping between function spaces. Traditional neural networks can only map between finite-dimensional euclidean space i.e., can only process finite-dimensional vectors or sets. Neural operator overcomes this limitation and is discretization invariant. Among the different parameterizations of the neural operator, Fourier Neural Operators (FNO) \cite{li2020fourier} have proven to be extremely effective in achieving state-of-the-art results in learning to solve different partial differential equations (PDE).  An FNO $\G$ can be expressed as 
$$\G: \{a: \D_A \rightarrow \Real^{d_a} \} \rightarrow \{u: \D_u \rightarrow \Real^{d_u} \} $$
where $a$ is the input function and $u$ is the output function. 
A neural Operator architecture consists of three principal components.
\begin{enumerate}
    \item Point-wise operator $P$ which transforms the input function $a$ to higher dimension spaces by mapping it to a function with higher co-domain dimension.
    \item A series of $n$ stacked non-linear neural integral operator layer $\{L_i\}_{i=1}^{i=n}$, where $n$ is a hyper-parameter of the model. Each layer $L_i$ is composed of a global kernel operator and a point-wise operator. The layer $L_i$ maps input function $v_i$ to the function $v_{i+1}$, which can be described as 
    $$ v_{i+1} =  L_i(v_i) = \sigma  (\int \kappa_i(x,y) v_i(x) d\mu(y) + W_i v_i) $$
    where $\mu$ is the measure over the domain of function $v_i$. The kernel integration is usually computed in the Fourier domain as $\F^{-1}( R_i \cdot \F (v_i))$, where $\F$ is the Fourier transform and $R_i$ is a complex-valued matrix representing kernel $\kappa_i$. And the $\sigma$ is a non-linearity of choice. In practice GELU \cite{hendrycks2016gaussian} is widely used \cite{li2020fourier, rahman2022u}.
    \item The series of integral operators are followed by a projection operator $Q$. It projects the output function of $L_n$ to match the desired output function $u$.
\end{enumerate}
Due to the ability to capture complex nonlinear dynamics, FNOs are being employed to solve complex real-world problems such as CO2 injection \cite{wen2022u}, weather modeling \cite{pathak2022fourcastnet}, and seismology \cite{yang2021seismic}. Recently proposed GANO \cite{rahman2022generative} introduces a generative adversarial approach to learn probability on function spaces generalizing GAN which is the first principled approach for generative models in function spaces. 

The problem of generating human motion for interactive activities conditioned by the motion of other interacting humans is not well-explored. Often in human-robot interaction, AR/VR, and VFX design, we need to generate the motion of digital humans depending on the motion of the real actors. To the best of our knowledge, PaCMO is the first work to address such an important problem and we do that by introducing a conditional generative neural operator architecture.
\section{Methodology}
\label{sec:method}
In this section, we provide the mathematical formulation of our problem. Then we formalize a general construction for the conditional generative adversarial neural operator. Next, we propose the ($\text{F}^2\text{ID}$) score, an adoption of Fréchet inception distance (FID) score for infinite dimensional vector (function) spaces.
\subsection{Problem Formulation}
We consider human motion as a sequence of skeletal poses. In skeletal representation with $J$ joints, the human pose or gesture can be represented as a vector $$ g = [x^1, y^1,z^1,x^2,y^2,z^2, \dots x^J,y^J,z^J ] \in \Real^{(3 \times J)}$$
where $(x^i,y^i,z^i)$ represents the location of $i$th joint in 3D euclidean space.
Let $\M$ be the motion space of humans such that for $m \in \M, m: t \rightarrow \{g \in \Real^{(3 \times J)}\}$, where $t \in \Real$ is the time. Then the probability space $(\M \times \M, \sigma(\M) \times \sigma(\M), \P_{\M \times \M})$ represents the probability space of motions for all dyadic human activities. In other words, for any $(m_1,m_2) \sim  \Prob_{\M \times \M}$, motion sequences $m_1$ and $m_2$ form a valid dyadic human activity when $\Prob_{\M \times \M} \in \P_{\M \times \M}$. Also we assume that $(m_1,m_2)$ and $(m_2,m_1)$ represents the same dyadic activity i.e., $m_1$ and $m_2$ are exchangeable.

Here we want to learn the conditional probability distribution $\Prob_{\M|\M}$ such that for any $m_2 \in \M$ and $ m_1 \sim \Prob_{\M|m_2}$, $m_1,m_2$ constitute a valid dyadic human activity. Note that we do not assume any particular time discretization or length of the motion sequence.

\subsection{Conditional Generative Adversarial Neural Operator (cGANO)}
Let $\U,\A,$ and $\C$ be Polish function spaces such that for any  
$$ \forall u \in \U, u: \D_\U \rightarrow \Real^{d_\U} , ~~\forall a \in \A, a: \D_\A \rightarrow \Real^{d_\A},$$
$$ \text{and~}\forall c \in \C, c: \D_\C \rightarrow \Real^{d_\C}$$
We define the space of operators $\bG$ (Generators) such that for any $\G \in \bG, \G: \A \times \C \rightarrow \U$ and we also define the space of functional $\bL$ (Discriminators) that for any $d \in \bL, d: \U \times \C \rightarrow \Real$.\\

Let the probability space $(\A, \sigma(A), \P_\A)$ is induced by Gaussian Random Field (GRF) and  $(\U \times \C, \sigma(\U) \times \sigma(\C), \P_{\U \times \C})$ denotes the probability space on the product space $\U \times \C$. We want to model the conditional probability $\Prob_{\U|\C}$ with the push-forward measure of $\Prob_\A$ under the map $\G$ given $\C$ i.e., $\bG_\C\sharp\Prob_\A$. Hence we define the Wasserstein distance between the two distributions 
as follows
\begin{equation}
    W(\Prob_{\U \times \C}, \Prob_{\C} \cdot \bG_\C\sharp\Prob_\A) = \sup_{\substack {d:d\in\bL \\ Lip(d)\leq 1}}\E_{\Prob_{\U \times \C}}[d]-\E_{\G\sharp\Prob_\A,\Prob_\C}[d]
    \label{eqn:wasser}
\end{equation}
where $\Prob_\C$ comes from the joint probability measure $\Prob_{\U \times \C}$. We aim to find $\G \in \bG$ that minimizes the Wasserstein distance between the true and generated probability measure. Assuming   $(\U \times \C)^*$ denotes the dual space of $\U \times \C$, we express the Lipshitz constraint as \cite{goodfellow2020generative, rahman2022generative}  
$$Lip(d)\leq 1 \iff \|\partial d\|_{(\U \times \C)^*} \le 1 $$

Following the construction of GANO, we can rewrite the objective without the constraint as

\begin{equation}
    \inf_{\G\in \bG} \sup_{d\in\bL}\E_{\Prob_{\U \times \C}}[d]-\E_{\Prob_{\C},\G\sharp\Prob_\A}[d] + \lambda \E_{\Prob'_{\U \times \C}}(\|\partial d\|_{(\U \times \C)^*}-1)^2
    \label{eqn:con_wass}
\end{equation}
where $\Prob'_{\U \times \C} := \lambda \Prob_\C \cdot \G\sharp\Prob_\A + (1 - \lambda) \Prob_{\U \times \C}$ given $\lambda \sim U(0,1)$.

Now we construct the operator on the product space of $\A \times \C$. Let's decompose $\bG$ as  $$\bG := \bG' \circ \bG_{con}$$ where $\G_{con}$ defines concatenation operator such that for any $\G_{con} \in \bG_{con}, \G_{con}: \A \times \C \rightarrow \V$. Here the function space $\V$ is defined such that for any 
$$v \in \V, v: (D_\A \times \D_\C) \rightarrow \Real^{d_\A+ d_\C} $$ and $$v(x \oplus y) = a(x) \oplus c(y) ~~\forall (x,y) \in (D_\A \times \D_\C).$$ 

Where $\oplus$ denotes the concatenation of functions and $\bG'$ is the space of operators mapping $\V$ to the target function space $\U$.

In many practical scenarios, we want to learn the conditional probability measure of the function space where the conditional variables are finite-dimensional (e.g., action labels). In other words, we want to model $\Prob_{\U|s}$ where $s \in \D_s$, where $\D_s$ is a finite-dimensional vector space. 

In such cases, we learn a mapping from $\D_s$ to $\C$. We define the mapping space $\mathcal{E}$ such that for any $E \in \mathcal{E}, E: \D_\C \times \D_s \rightarrow \Real^{d_\C}$. For any $s \in D_s$ the map $E$ maps it to $c_s \in \C$ such that 
$$c_s(y) = E(y \oplus s) ~~~~\forall y \in \D_\C$$

In this work, we use U-NO \cite{rahman2022u}, a U-shaped Fourier neural operator architecture for the generator (see Fig. \ref{fig:gen}). For the discriminator, we also use the U-NO architecture followed by a functional layer (see Fig. \ref{fig:dis}). A functional layer maps functions to real numbers which can be described as following
$$ r = \int_{D_v} \kappa_f(x) v_n(x) dx$$
where $v: \D_v \rightarrow \Real$ is the input function, $r$ is a real number, and $\kappa_f: \D_v \rightarrow \Real$ is a learnable function in the functional layer implemented by a fully connected neural network. 

For conditional human motion generation, the function space $\C$ and $\U$ is the 3D human motion space i.e., $\C = \U = \M$, i.e., for any operator $\G \in \bG$  
$$\G: \A \times \M \rightarrow \M$$
where for any Gaussian process $a \in \A$, $m_2 \in \M$, and $m_1 = \G(a,m_2)$, $(m_1,m_2)$ constitutes a valid dyadic human motion for two person.

\subsection{Construction of ($\text{F}^2\text{ID}$)}
Quantitative evaluation of high-dimensional generative models is an open problem. Among different proposed metrics, Fréchet Inception Distance or FID \cite{heusel2017gans} is well used. Though it was first introduced to evaluate generative models for images, it has been adapted for other generative tasks (e.g., motion, voice). FID measures the Wasserstein-2 distance between the distribution of inception features of the generated and real images. For other modalities, features from the real and generated data are extracted using suitable models. To get a closed-form solution of Wasserstein-2 distance, the extracted features from real and generated data are modeled by two multivariate  Gaussian distributions. Let $(\mu, \Sigma)$, $(\mu', \Sigma')$ be the mean and covariance of real and generated data respectively. Then the FID score is calculated as 
%
%
\begin{align*}
 FID &= W_2^2({\mathcal {N}}(\mu ,\Sigma ),{\mathcal {N}}(\mu ',\Sigma '))^{2}\\
&=\lVert \mu -\mu '\rVert _{2}^{2}+\operatorname {tr} \left(\Sigma +\Sigma '-2\left(\Sigma^{\frac12} \cdot \Sigma '\cdot \Sigma^{\frac12}\right)^{\frac {1}{2}}\right).
\end{align*}
But multi-variate Gaussian distributions are not suitable to model data (or features) from infinite-dimensional function spaces or processes (both stochastic and deterministic). Therefore, the FID score is not a suitable metric for evaluating generative models in function spaces.

To address this problem, we introduce Functional FID ($\text{F}^2\text{ID}$) scores. Here, we model the extracted feature by the Gaussian process. A Gaussian process, $GP(m,k)$,  is defined by a mean function $m$ and a co-variance function $k$. Wasserstein-2 distance between Gaussian processes is well defined and has a closed-form solution under the assumption that they are indexed over a compact domain $X \subset \Real^n$ with a metric $d(.,.)$, $m \in L^2(x)$, and $k \in L^2(X \times X)$ \cite{mallasto2017gp}. If the features extracted from real and generated data are modeled by two Gaussian processes $GP(m_1,k_1)$ and $GP(m_2,k_2)$ with associated covariance operator $K_1$ and $K_2$ respectively. Then the $\text{F}^2\text{ID}$ score is defined as 
\begin{equation}
    \text{F}^2\text{ID} = d_2(m_1,m_2) + Tr(K_1+K_2 - 2(K_1^{\frac12} \cdot K_2 \cdot K_1^{\frac12}))^{\frac12}
    \label{eqn:cFID}
\end{equation}
where for any operator $T$, $Tr(T)$ is the trace of the operator $T$ and we take $d(.,.)$ as the metric induced by $L^2$ norm.
 In this work,  we define $GP$ over time $t$. Given the features from the generated and real motions, we model them by the Gaussian processes $GP_1(m_1,k_1)$ and $GP_2(m_2,k_2)$ respectively using moment matching. We evaluate mean and covariance functions over a finite discretization of the domain.  Finally, we apply the Eqn.\ref{eqn:cFID} to calculate $\text{F}^2\text{ID}$ (details in supplementary). Finite resolution computation of Wasserstein-2 distance is proven to converge to true Wasserstein-2 distance with the increase of discretization resolution \cite[Theorem 8]{mallasto2017gp}.

\begin{figure*}[h]
    \centering
    \includegraphics[width=0.95\textwidth,trim={1.5cm 11cm 2.47cm 6cm},clip]{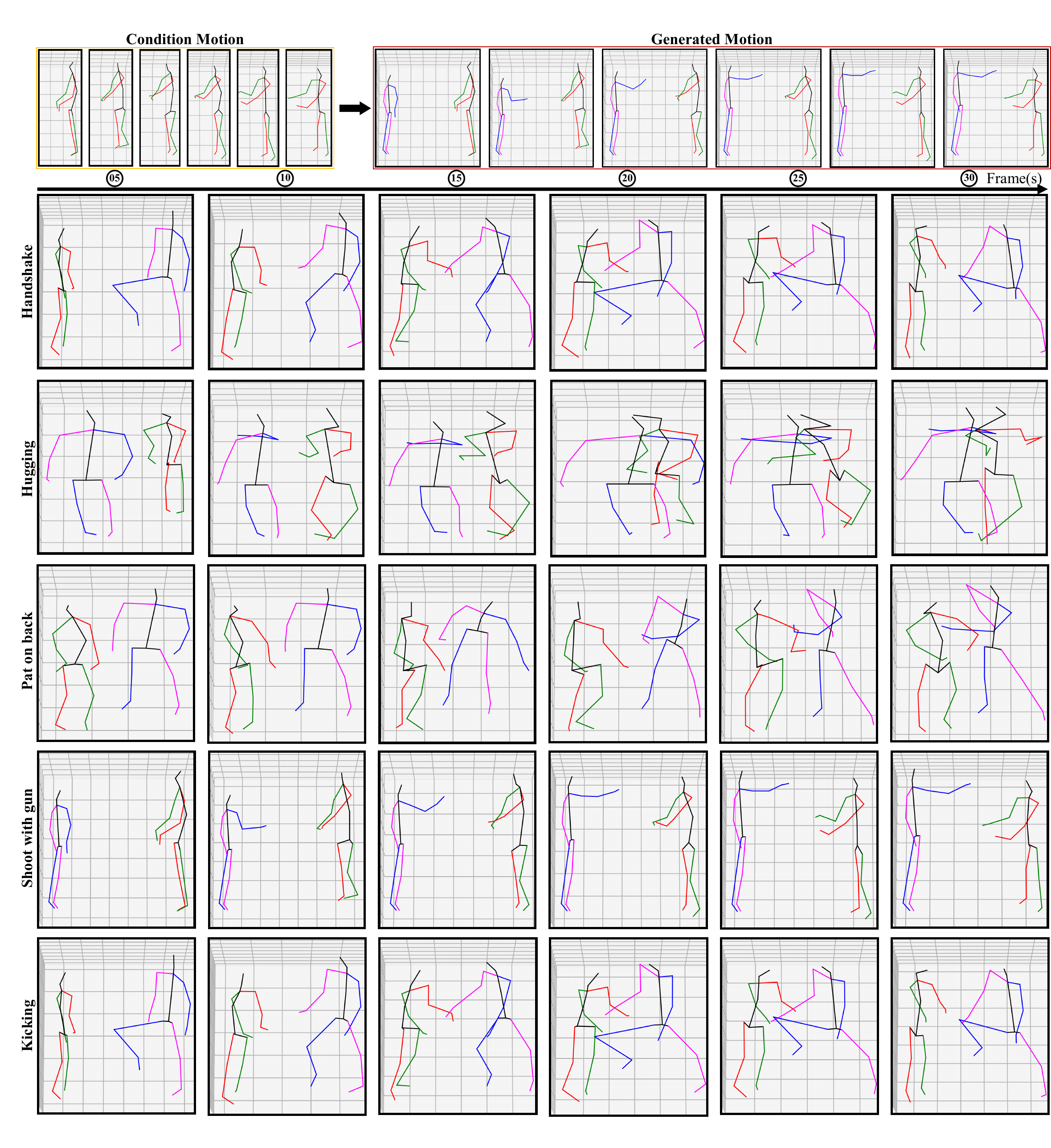}
    \caption{Samples of conditional generation of human motion by PaCMO. The top row shows the input condition motion (Red and Green) on the left and the complete motion on the right by combining the generated motion (in Magenta and Blue) and input motion (Red and Green). The following rows only show the complete motion combining the condition motions and generated motions from PaCMO. Different task labels are indicated on the left. Though PaCMO does not take action labels or descriptions as input, we can observe that PaCMO generated motions interact with the input motions depending on the action.  }
    \label{fig:result_sample}
\end{figure*}

\section{Experiments}
\label{sec:experiments}
\begin{table*}[]
\begin{tabular}{ccccccc}
\hline
\multicolumn{7}{c}{NTU-RGB+D}                                                                                                                                                                                                \\ \hline
\multicolumn{1}{c|}{Methods}      & \multicolumn{1}{c|}{$\text{F}^2\text{ID}$ $\downarrow$}  & \multicolumn{1}{c|}{Diversity Score $\rightarrow$} & \multicolumn{1}{c|}{MMD-A $\downarrow$}  & \multicolumn{1}{c|}{MMD-S$\downarrow$}  & \multicolumn{1}{c|}{APE (root joint)$\downarrow$} & AVE(root joint)$\downarrow$ \\ \hline
\multicolumn{1}{c|}{Real Motion}  & \multicolumn{1}{c|}{0.001} & \multicolumn{1}{c|}{1.90}             & \multicolumn{1}{c|}{0}      & \multicolumn{1}{c|}{0}      & \multicolumn{1}{c|}{0}                & 0               \\ \hline
\multicolumn{1}{c|}{PaCMO}        & \multicolumn{1}{c|}{0.002} & \multicolumn{1}{c|}{1.91}             & \multicolumn{1}{c|}{5.8e-3} & \multicolumn{1}{c|}{3.8e-3} & \multicolumn{1}{c|}{0.03}           & 0.23          \\ \hline
\multicolumn{1}{c|}{GRU baseline} & \multicolumn{1}{c|}{0.351} & \multicolumn{1}{c|}{1.92}             & \multicolumn{1}{c|}{7.4e-3} & \multicolumn{1}{c|}{4.4e-3} & \multicolumn{1}{c|}{0.05}           & 0.35          \\ \hline
\end{tabular}
\caption{Performance NTU-RGB+D 120 dataset. $(\downarrow)$ denotes the lower the score the better and $(\rightarrow)$ denotes the closer to the real dataset the better. Each metric is calculated five times and the \textbf{mean} is reported.}
\label{ref:tab_ntu}
\end{table*}

\begin{table*}[]
\begin{tabular}{ccccccc}
\hline
\multicolumn{7}{c}{DuetDance}                                                                                                                                                                                    \\ \hline
\multicolumn{1}{c|}{Methods}      & \multicolumn{1}{c|}{$\text{F}^2\text{ID}$ $\downarrow$}    & \multicolumn{1}{c|}{Diversity Score$\rightarrow$} & \multicolumn{1}{c|}{MMD-A $\downarrow$}   & \multicolumn{1}{c|}{MMD-S$\downarrow$}   & \multicolumn{1}{c|}{APE (root joint)$\downarrow$} & AVE(root joint)$\downarrow$ \\ \hline
\multicolumn{1}{c|}{Real Motion}  & \multicolumn{1}{c|}{1.6e-4} & \multicolumn{1}{c|}{0.261}     & \multicolumn{1}{c|}{0}       & \multicolumn{1}{c|}{0}       & \multicolumn{1}{c|}{0}           & 0          \\ \hline
\multicolumn{1}{c|}{PaCMO}        & \multicolumn{1}{c|}{4.0e-4}  & \multicolumn{1}{c|}{0.264}      & \multicolumn{1}{c|}{6.7e-3}  & \multicolumn{1}{c|}{6.67e-3} & \multicolumn{1}{c|}{0.01}      & 0.01    \\ \hline
\multicolumn{1}{c|}{GRU baseline} & \multicolumn{1}{c|}{494.9e-4} & \multicolumn{1}{c|}{0.190}     & \multicolumn{1}{c|}{23.36e-3} & \multicolumn{1}{c|}{23.35e-3} & \multicolumn{1}{c|}{0.04}      & 0.08    \\ \hline
\end{tabular}
\caption{Performance DuetDance dataset. $(\downarrow)$ denotes the lower the score the better and $(\rightarrow)$ denotes the closer to the real dataset the better.  Each metric is calculated five times and the \textbf{mean} is reported.}
\label{ref:tab_duet}
\end{table*}

\subsection{Datasets}
For the purpose of our work, we need an interactive dyadic human activity dataset with skeletal pose annotation. Unfortunately, such publicly available datasets are scarce. We use two existing datasets:  NTU-RGB+D 120\cite{liu2019ntu} and the DuetDance \cite{kundu2020duet} datasets.
\begin{itemize}
    \item \textbf{NTU-RGB+D} 120 is the extended version of original NTU-RGB+D \cite{shahroudy2016ntuo} dataset. It contains 120 action types (such as handshake, pushing) of 106 subjects. The provided skeleton data contains 3D locations of 25 major body joints. For our purpose, we only considered actions with two persons resulting in 26K sequences (motion pairs). Though we do not clip any motion to restrict the length, we are not modeling the movement of the finger, as it requires finer precision noise-free data \cite{taheri2022goal} and we exclude two joints from each hand representing the tip of the finger and thumb. 
    \item \textbf{DuetDance} The dataset is created from complex dance poses such as Cha-cha, Jive, Rumba, Salsa, and Samba from the curated video tutorial on Youtube. Due to the nature of the actions, the motion encompasses long-term correlations between physical motions. From the curated videos the skeletal positions are calculated using LCRNet++ \cite{rogez2019lcr}. The skeletal data contain 3D location of 15 major body joints. The dataset consists of one very long sequence ($\sim$ 48K time steps) for each of the dance types. So to create more training instances, we divide one long sequence into smaller sequences non-overlapping sequences.  
\end{itemize}

\subsection{Hyperparameters}
We use U-shaped neural operator architecture \cite{rahman2022u} in both the generator and discriminator network similar to GANO \cite{rahman2022generative}. We use seven stacked integral operator layers with GELU as non-linearity (see Fig. \ref{fig:gen} and \ref{fig:dis}). We divide each dataset into training and evaluation sets randomly at a ratio of $8:2$.  For each dataset, we train our model for 200 epochs with a single NVIDIA A100  GPU on the training set. We used an ADAM optimizer with an initial learning rate of $10^{-4}$ which is halved after every 50 epochs. We train our model on the skeletal data represented as joint locations in 3D space. While training we do not limit the length of the motion sequence by clipping.  
To maintain the symmetry of the human body, We also add additional constraints on the generator by penalizing asymmetric limb generations i.e., restricting the generator to produce left and right legs of the same length with soft constraints. Empirically we have found that these soft constraints increase the convergence rate and the realism of the generated motions.

\subsection{Experiment Results}

\subsubsection{ Evaluation Metrics}
\begin{itemize}
    \item \textbf{$\text{F}^2\text{ID}$}: We used our proposed $\text{F}^2\text{ID}$ score as one of the the evaluation metric. To extract features from the motion sequences, we have trained an auto-encoder neural operator to extract features from human motion. The auto-encoder operator can be defined as $\G: (\M \times \M) \rightarrow \M \times \M$. The auto-encoder also is composed of seven non-linear integral operator layers and the output of the bottleneck layer is taken as a feature. We extract features from randomly sampled (with replacement) 1000 generated and real dyadic human motions and calculate the $\text{F}^2\text{ID}$. The process is repeated five times and the mean $\text{F}^2\text{ID}$ score is reported.
    \item \textbf{Diversity Score}: Diversity score represents the variance in the generated data. We randomly sample (with replacement) 1000 motion from the generator and extract the features using the auto-encoder. We divide the features into two sets $S$ and $S'$ each containing 500 motions. Then the diversity score is calculated as
     \vspace{-5pt}
    $$ \text{Diversity Score } = \frac{1}{5e2} \sum_{i=1}^{5e2} L_2(S_i,S'_i)^2 $$
    where $L_2(.,.)$ is the $L_2$ norm. The average diversity score of five repeated calculations is reported.
    
    \item \textbf{Maximum Mean Discrepancy (MMD)} Maximum Mean Discrepancy measures the distance between two distributions and is used to measure the discrepancy between generated and real data \cite{battan2021glocalnet, yu2020structure, tolstikhin2016minimax}. Following the construction in \cite{maheshwari2022mugl}, we also used the kernel trick to calculate the MMD. We also present two MMD-based metric MMD-A which measures the dissimilarity per time step basis and MMD-S which measure the dissimilarity based on the whole sequence. We use the official implementation of the work MUGL \cite{maheshwari2022mugl}. 
    \item \textbf{Average Position Error (APR)} Average position error for joint $j$ between generated motion $\{m_i'\}_{i=0}^n$ and the corresponding ground truth real motion $\{m_i\}_{i=0}^n$is defined as 
    $$ APE(j) = \frac{1}{N} \sum_{i=0}^{n} \frac{1}{T_i} \sum_{t=1}^{T_i} ||m'^j_i(t) - m^j_i(t)||_2$$
    where $m^j_i(t)$ is the position of $jth$ joint of motion $m_i$ at time $t$ and $T_i$ is length of motion $m_i$
    \item \textbf{Average Variance Error (AVE)} Following the above notation, AVE can be defined as 
    $$AVE(j) = \frac{1}{N} \sum_{i=0}^{n}  || \sigma(m'^j_i) - \sigma(m^j_i)||_2$$
    where $\sigma(m^j_i)$ is the variance of joint $j$ over time. We calculate both APR and AVE without any normalization of the data.
\end{itemize}
 It is pointed out in recent studies \cite{maheshwari2022mugl} that feature-based metrics (e.g., accuracy) are not reliable to evaluate action sequences. As a result, we kept both feature-based (FID, Diversity score) metrics and metrics calculated directly from the generated and real data (MMD, AVE, APE) as our evaluation metrics. We report the mean value for every metric over five repeated calculations on the randomly sampled data from the evaluation set.
 \vspace{-7pt}
\subsubsection{Baseline}
 \vspace{-5pt}
The problem of partner-conditioned human motion generation in function spaces is new and we do not have any method to compare with. Previously, LSTM-based models are used to generate body motion from audio \cite{shlizerman2018audio}. Here we also design a GRU base model to predict our desired human motion from the motion of their partner in the action. The model consists of five stacked GRU layers with a $200$ dimensional hidden state. This is followed by three layered fully connected neural networks with GELU activation. This type of sequence-to-sequence (seq-to-seq) architecture has been used successfully in many complex real-world problems such as language translation, weather prediction \cite{sehovac2020deep, sehovac2019forecasting,mangal2019lstm,shewalkar2019performance}. So, this model serves as an appropriate deterministic baseline for this problem.   

\begin{table}[h]
\begin{tabular}{c|c|c}
\hline
                                                                & Real data & \begin{tabular}[c]{@{}c@{}}PaCMO Generated\\ data\end{tabular} \\ \hline
\begin{tabular}[c]{@{}c@{}}Perceived as\\ Realistic\end{tabular}   & 0.81      & 0.75                                                           \\ \hline
\begin{tabular}[c]{@{}c@{}}Perceived as\\ Unrealistic\end{tabular} & 0.19      & 0.25                                                           \\ \hline

\end{tabular}
\caption{Result of the survey of real and PaCMO generated motion.}
\label{tab:survey}
 \vspace{-10pt}
\end{table}

 \vspace{-7pt}
\subsubsection{Results}
 \vspace{-5pt}
We present the performance of PaCMO on the DuetDance dataset shown in Table \ref{ref:tab_duet} and on the NTU-RGD+D 120 dataset shown in Table \ref{ref:tab_ntu}. $\text{F}^2\text{ID}$ is most important in evaluating the quality of the generated motions. In both the datasets, PaCMO achieves lower $\text{F}^2\text{ID}$ vastly outperforming the baseline and the diversity score of PaCMO is closer to the real dataset. Unlike motion generation from the action label where the input condition is just a fixed label, here the input condition is the motion of the partner for a dyadic activity and contains much variability even within a single action category. As a result, both GRU base seq-to-seq models and PaCMO achieved a diversity score close to the real dataset. For feature-free evaluation metrics (MMD, APE, AVE), PaCMO achieves excellent results outperforming the baseline. This shows that the motion generated by PaCMO is very realistic and close to the real dataset. 
 \vspace{-8pt}
\subsubsection{Perceptual Evaluation}
 \vspace{-5pt}
To bolster the finding in the previous section, we also conduct a visual perceptual evaluation using of 50 web-based participants. In this study, we provide the participants with both real and PaCMO-generated motions and ask them to label them as real or fake (computer generated). Each participant is asked to label 14 different randomly selected motions from both real and generated data. The result of the study is shown in Table \ref{tab:survey}. We can observe that our generated motion is labeled as real at a very high rate ($76\%$), which makes it suitable for real applications such as AR/VR and animation.
For qualitative evaluation, we present a few motions generated by PaCMO along the condition motion (see Fig. \ref{fig:result_sample}) and we observe that the generated motions are realistic and in sync with the given motion of the partner as a condition.

 \vspace{-15pt}
 \section{Conclusion, Limitations, and Future}
\label{sec:conclusion}
This work addresses an important yet under-explored problem of partner-dependent human motion generation in dyadic human activities. And as a solution, we propose PaCMO, a novel generative neural operator architecture. Unlike existing conditional generative approaches, PaCMO is resolution invariant, able to model global long-term dependencies, and also does not depend on any particular representation of human poses. We also propose $\text{F}^2\text{ID}$ as a suitable evaluation metric for generative models for infinite dimensional spaces. Our work is the first step in generating continuous conditional human motion generation. In the future, we plan to extend the work from dyadic activities to multi-person group activities. Also, we aim to model complected human motion such as grasping and emotions.

{\small
\bibliographystyle{ieee_fullname}
\bibliography{egbib}
}

\appendix
\newpage
\newpage
\section{Supplementary Material}
\subsection{Calculation of \FID}
In this section we will discuss the approximation of Wasserstein-2 distance for Gaussian processes and \FID.
\begin{algorithm}[ht]
\caption{Calculation of \FID}\label{alg:FID}
\KwData{Features of real data $F_t$, features of generated data $F_g$, the grid size $r$ }
\KwResult{\FID~score}
model $F_t$ by $GP(m_1,k_1)$\;
model  $F_g$ by $GP(m_2,k_2)$\;
g[] $\leftarrow$ $r\text{ evenly spaced number on the interval }[0,1]$\;
Initialize matrix $M_t$ and $M_g$ of size $r \times r$\;
\For{i in $\{1,2, \dots r \}$}{
\For{ j in $\{1,2, \dots r \}$}
{$M_t[i][j]$ $\leftarrow$ $\frac{1}{r} k_1(g[i],g[j])$ \;
$M_g[i][j]$ $\leftarrow$ $\frac{1}{r} k_2(g[i],g[j])$ \;
}}
L $\leftarrow$ $M_t+M_g - 2(M_t^\frac12 M_g M_t^\frac12)^\frac12 $\;
\FID = $\frac1r \sum_{i=1}^{r} (m_1(g[i]) - m_2(g[i]))^2 +  \sum_{i=1}^{r}L[g[i]][g[i]] $\;
\end{algorithm}

Let $GP(m,k)$ be a Gaussian process where the mean function $m \in L^2(X)$, and the covariance function $k \in L^2(X \times X)$. For the simplicity of calculation, without any loss of generality we will assume that $X$ is a compact subset of $\Real^d$.  We also assume that  $\lambda(X) = 1$ with Lebesgue measure $\lambda$ on $R^d$. We assume  that the function in  $L^2(X \times X)$ are discretized uniformly at the finite set of point $x_i \in X_i$. 

For numerical feasibility, we choose only a finite set of abnormal basis $\{e_1,e_2,..e_r\}$ for the function spaces $L^2(X)$. The domain $X$ is discretized uniformly with a  grid size of $r$ such that  $\lambda(X_i) = \frac1r ~~\forall i  \in \{1,2, \dots r\}$ and $\bigcup_{i= 0}^{r}X_i = X$. 

The basis function $e_i$ is defined to be equal to $\frac{1}{c}\sqrt{r}$ at the point $x_i$ and $0$ outsize the domain $X_i$. Such choice constants maintains the basis functions at unit norm.  The $c$ is the universal constant depends only the particular choice of $e_i$ \cite{zienkiewicz2005finite}, we assume it to be $\approx 1$ for our calculation.

The covariance operator $K$ associated with covariance function $k$ is defined as 
\begin{equation}
    [ K \phi](x) = \int_X k(x,s) \phi(s) ds
    \label{eqn:cov_op}
\end{equation}
Applying the operator $K$ on the the basis function $e_i$, we get

\begin{align*}
    [ K e_i](x) &= \int_X k(x,s) e_i(s) ds\\
                &= \int_{X_i} k(x,s) ds\\
                & \approx \frac{1}{\sqrt{r}} k(x,x_i) ~~~ [\text{where }x_i \in X_i]
\end{align*}
Here we assume that for a fixed $x$ the covariance function $k(x,s)$ is constant $\forall s \in X_i$.

The trace of the operator $K$ can be defined as $$Tr(K) = \sum_{i=1}^r \langle Ke_i , e_i \rangle$$
Now,

\begin{align}
\begin{split}
    \langle Ke_i , e_i \rangle &= \int_X [Ke_i](x) e_i(x) dx\\
    &= \int_X \frac{1}{\sqrt{r}} k(x,x_i) e_i(x) dx\\
    & \approx \int_{X_i} k(x,x_i)dx \\
    & = \frac{1}{r} k(x_i,x_i)
\end{split}
\label{eqn:dot_pord}
\end{align}
where we assumes that for a fixed $x_i$, $k(x,x_i)$ is constant $\forall x \in X_i$. Now the trace can be written as 
$$Tr(K) \approx \frac{1}{r}\sum_{i=1}^r k(x_i,x_i)$$
where $x_i \in X_i$.
Assuming a finite set of standard basis we can construct covariance matrix $M_K$ associated with the covariance operator $K$ defined as follows
$$M_K[i,j] = \langle Ke_i , e_j \rangle$$
where $i,j \in \{1,2, \dots r\}$. Following the same procedure in Eqn. \ref{eqn:dot_pord}, we can show that  $M_K[i,j] \approx \frac{1}{r} k(x_i,x_j)$ where $x_i \in X_i$ and $x_i \in X_j$.

We present a pseudo-code for calculating \FID~ in Algorithm \ref{alg:FID}. We assume that the feature functions are defined on the interval $[0,1]$. The $F_t = \{f_t^1,f_t^2, ...,f_t^n\}$ and $F_g = \{f_g^1,f_g^2, ...,f_g^{n}\}$ are the set of $n$ feature functions of the real and generated motion respectively where $f: [0,1] \rightarrow \Real$. The feature functions are extracted from the bottleneck layer of the auto-encoder neural operator. The auto-encoder neural operator follows the architecture of PaCMO generator without the skip connections.
Each of the feature functions $f^i$ for $i \in \{1,2,...,n\}$ for both real and generated is represented using a finite set of basis as following
$$
f^i(x) = \sum_{j=1}^{r^i} F^i[j] e^i_j
$$
where $r^i$ is resolution (discretization grid size) of the feature function $f^i$ which depends on the input resolution feed to the auto-encoder neural operator. The $e^i_j$ s are the corresponding basis function which depend on $r^i$ and $F^i$ s are the coefficients.
We model the the features from real data by Gaussian process $GP(m_1,k_1)$ where
$$
m_1(x) = \E[f_t] \approx \frac1n \sum_i^n f_t^i(x)
$$
and
\begin{align*}
   k_1(x,x') &= \E[\big(f_t(x)-m_1(x)\big)\big(f_t(x') - m_1(x')\big)] \\
   & \approx \frac1n \sum_i^n (f^i_t(x) - m_1(x)) (f^i_t(x') - m_1(x'))
\end{align*}

Here the expectations are taken over the distributions of feature functions for real data and approximated by the the set $F_t$. In the same way we model the feature functions of the generated data by $GP(m_2,k_2)$. To approximate the 2-Wasserstein metric between these Gaussian processes we fix a resolution $r$. And the matrix operations are computed using Scipy \cite{virtanen2020scipy}.

\subsection{More Qualitative Results}

\begin{figure*}[h]
    \centering
    \includegraphics[width=0.75\textwidth,trim={0cm 5cm 8cm 0cm},clip]{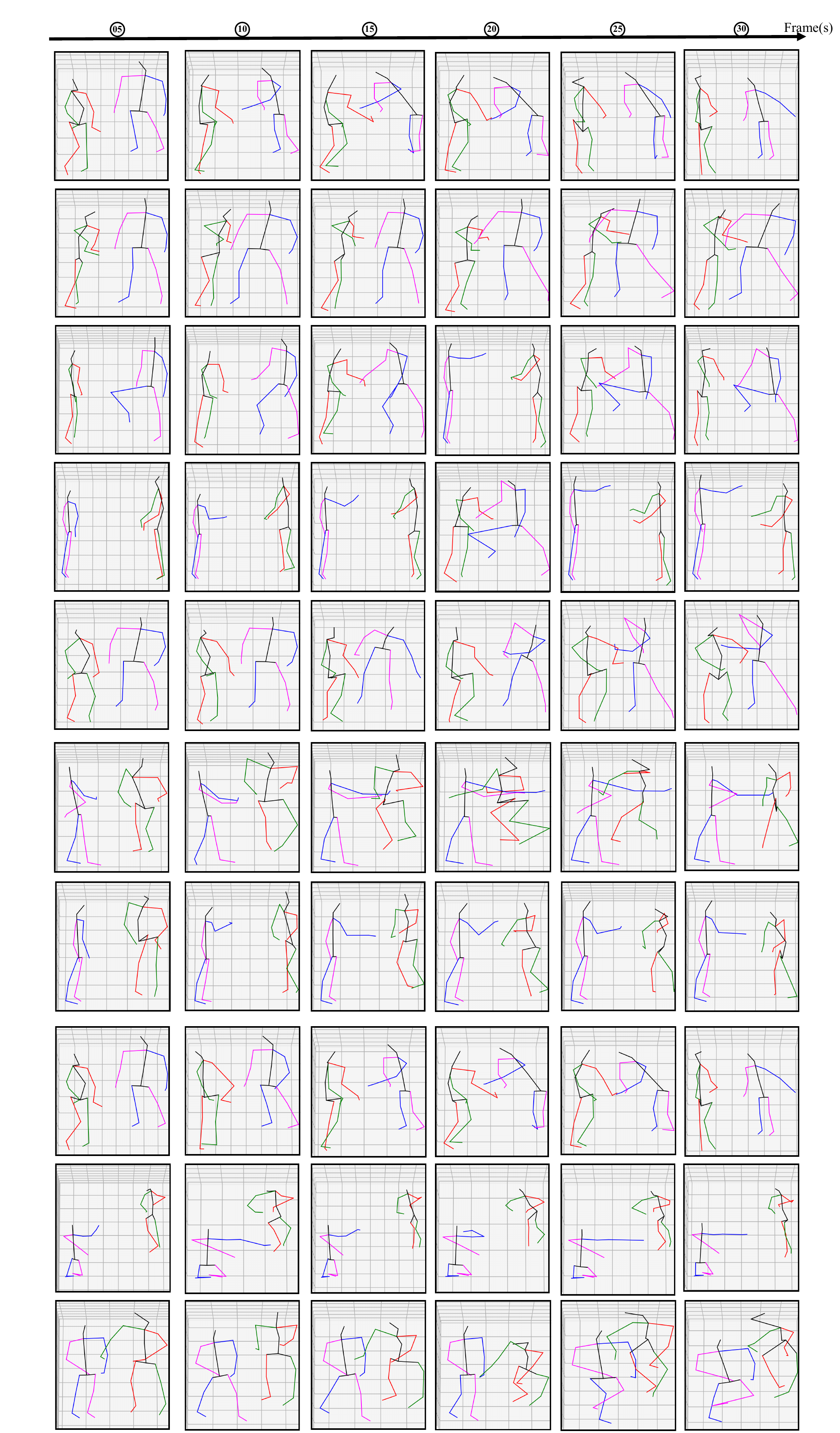}
    \caption{Each rows represents a complete dyadic motion with the condition  motion (in Red and Green) and motion generated by PaCMO (in Magenta and Blue). }
    \label{fig:more_results}
\end{figure*}

\end{document}